\documentclass{article}
\usepackage{ijcai25}
\usepackage{times}
\usepackage{soul}
\usepackage{url}
\usepackage[hidelinks]{hyperref}
\usepackage[utf8]{inputenc}
\usepackage[small]{caption}
\usepackage{graphicx}
\usepackage{amsmath}
\usepackage{amsthm}
\usepackage{booktabs}
\usepackage{algorithm}
\usepackage{algorithmic}
\usepackage[switch]{lineno}
\usepackage{dashrule}
\usepackage{amssymb}
\usepackage{comment}

\title{``Actionable Help" in Crises: A Novel Dataset and Resource-Efficient Models for Identifying Request and Offer Social Media Posts}

\author{
Rabindra Lamsal$^1$
\and
Maria Rodriguez Read$^1$\and
Shanika Karunasekera$^{1}$\And
Muhammad Imran$^2$\\
\affiliations
$^1$The University of Melbourne, Australia\\
$^2$Qatar Computing Research Institute, Qatar\\
\emails
\{r.lamsal, maria.read, karus\}@unimelb.edu.au,
\{mimran\}@hbku.edu.qa
}

\begin{document}

\maketitle

\begin{abstract}
During crises, social media serves as a crucial coordination tool, but the vast influx of posts~---~from ``actionable" requests and offers to generic content like emotional support, behavioural guidance, or outdated information~---~complicates effective classification. Although generative LLMs (Large Language Models) can address this issue with few-shot classification, their high computational demands limit real-time crisis response. While fine-tuning encoder-only models (e.g., BERT) is a popular choice, these models still exhibit higher inference times in resource-constrained environments. Moreover, although distilled variants (e.g., DistilBERT) exist, they are not tailored for the crisis domain. To address these challenges, we make two key contributions. First, we present \texttt{CrisisHelpOffer}, a novel dataset of 101k tweets collaboratively labelled by generative LLMs and validated by humans, specifically designed to distinguish actionable content from noise. Second, we introduce the first crisis-specific \texttt{mini models} optimized for deployment in resource-constrained settings. Across 13 crisis classification tasks, our mini models surpass BERT$_{\text{base}}$\footnote{Our models often outperform or match the performance of base architectures of RoBERTa, MPNet, and BERTweet.}, offering higher accuracy with significantly smaller sizes and faster speeds. The Medium model is 47\% smaller with 3.8\% higher accuracy at 3.5× speed, the Small model is 68\% smaller with a 1.8\% accuracy gain at 7.7× speed, and the Tiny model, 83\% smaller, matches BERT’s accuracy at 18.6× speed. All models outperform existing distilled variants, setting new benchmarks. Finally, as a case study, we analyze social media posts from a global crisis to explore help-seeking and assistance-offering behaviours in selected developing and developed countries.
\end{abstract}

\section{Introduction}
Social media has become essential for rapid communication and coordination during natural disasters or public health emergencies \cite{imran2015processing,lamsal2022socially}. People use these platforms to stay informed, check on loved ones, share their safety status, and request or offer assistance. During emergencies, individuals or organizations, either for themselves or on behalf of others, often post requests or offers of assistance, covering needs such as medical supplies, blood donations, clothes, food, water, and other essential resources \cite{purohit2014emergency}. By accurately identifying and matching these requests and offers, social media can be harnessed as a powerful tool to enhance crisis response and resource allocation, leading to more effective relief efforts.

During a crisis, conversations can grow exponentially, sometimes reaching hundreds of thousands per hour. For example, during the early stages of the COVID-19 pandemic, Twitter (now ``X") recorded over 2 million pandemic-related tweets every hour\footnote{\url{https://blog.x.com/engineering/en_us/topics/insights/2021/how-we-built-a-data-stream-to-assist-with-covid-19-research}}. During such surge of information, accurately identifying actionable posts~---~explicit requests for help or offers of assistance~---~becomes critical. Multiple studies in crisis informatics have attempted to identify requests and offers from social media discourse \cite{purohit2014emergency,nazer2016finding,devaraj2020machine,ullah2021rweetminer,lamsal2024crema}, but struggle to distinguish actionable posts from non-actionable ones. Posts such as emotional support (``Pray for the victims..."), behavioural guidance (``Please wear a mask."), or outdated information (``We were offering [something] earlier for free, but not now.") are frequently misclassified. Additionally, promotional content disguised as help (``Please shop online with us by clicking the link in our bio...") complicates classification further. One major reason existing approaches struggle with such nuanced cases is their lack of generalizability, which largely stems from the lack of a comprehensive dataset. Without high-quality and diverse data, models fail to capture subtle patterns critical for identifying actionable posts.

Our findings show that while generative LLMs can achieve high accuracy in identifying actionable posts in a few-shot setting, their high computational costs limit their applicability for real-time crisis response. Deploying LLMs in a real-time social media data pipeline, where thousands of posts may arrive continuously, is therefore impractical. A more practical approach is to fine-tune masked language models (MLMs) such as BERT \cite{devlin2018bert} and RoBERTa \cite{liu2019roberta}. While more feasible, these models still exhibit high inference times in resource-constrained environments due to their large parameter sizes. Research on parameter redundancy in pre-trained models \cite{voita2019analyzing,kovaleva2019revealing} highlights opportunities for smaller, more efficient models. However, existing mini MLMs \cite{sun2019patient,turc2019well,jiao2019tinybert,sanh2020distilbertdistilledversionbert} remain general-purpose, leaving a gap in domain-specific models and benchmarks for crisis informatics.

To address these challenges, this study makes two key contributions to the existing crisis informatics literature:

First, we introduce CrisisHelpOffer, a large-scale, high-quality labelled dataset designed to identify actionable requests and offers shared on social media during crises. The dataset is created using an ensemble of generative LLMs for the initial labelling, with human annotators validating the data to ensure consistency and reliability. By focusing on explicit, actionable content, CrisisHelpOffer provides a robust foundation for training classification models that can effectively filter noise and prioritize posts relevant to crisis response.

Second, we propose a set of smaller, crisis-domain-specific models with varying architectures and parameter counts. These \textit{mini models} balance computational efficiency and classification accuracy, making them suitable for real-time processing in crisis scenarios. Using a corpus of $\approx$440 million crisis-related tweets, we train these models to mimic the embedding space of a crisis-domain-specific base model. We evaluate their performance against widely used MLMs in crisis informatics~---~BERT, RoBERTa, MPNet, and BERTweet~---~across 13 crisis classification tasks, including CrisisHelpOffer. Additionally, we also compare our models with distilled counterparts such as DistilBERT, BERT$_{\text{medium/small/mini}}$, and TinyBERT, demonstrating their efficiency.

In summary, this study contributes the following:

\begin{itemize}
\item We present CrisisHelpOffer\footnote{The dataset will be provided upon request.}, a dataset of 101k tweets collectively labelled by multiple generative LLMs and validated by humans. To the best of our knowledge, this is the first dataset focused on identifying ``actionable" requests and offers shared on social media during crises.
\item We introduce\footnote{\texttt{https://huggingface.co/crisistransformers}} the first set of mini models for processing crisis-related social media texts. When evaluated on 13 different crisis classification tasks, our mini models often outperform or match the performance of base architectures of BERT, RoBERTa, MPNet, and BERTweet, and outperform commonly used distilled variants.
\item As a case study, we analyze a global crisis event to explore help-request and assistance-offering behaviours in selected countries, both overall and across specific resource types: money, volunteers, shelter, and food.
\end{itemize}

\section{Related Work}
\subsection{Identifying Request and Offers}
Multiple efforts have been made to identify \cite{purohit2014emergency,nazer2016finding,devaraj2020machine,ullah2021rweetminer,lamsal2024crema} and match \cite{purohit2014emergency,dutt2019utilizing,lamsal2024crema} requests and offers shared on social media during crises. Early work by \cite{purohit2014emergency} used regular expressions and labelled data to classify requests and offers via two sequential Random Forest classifiers. \cite{nazer2016finding} improved classification by incorporating topic models, URLs, hashtags, and user metadata. \cite{ullah2021rweetminer} further refined this approach using rule-based features.

Dense vector representations have been used for similar classification tasks. \cite{devaraj2020machine} employed GloVe vectors \cite{pennington2014glove}, n-grams, and POS tags to classify tweets as ``urgent" or ``not urgent," while \cite{he2017signals} combined n-grams with word2vec \cite{Mikolov2013EfficientEO} and trained an XGBoost classifier to detect ``logistical information" tweets. Recently, transformer-based models \cite{vaswani2017attention} have shown strong performance \cite{zhou2022victimfinder,lamsal2024crema}, with CrisisTransformers \cite{lamsal2024crisistransformers}~---~trained on over 15 billion tokens from 30+ crisis events~---~setting the current state-of-the-art.

The dataset from \cite{purohit2014emergency} is seminal in this field. A recent study \cite{lamsal2024crema}, however, reported that only 41\% of tweets classified as requests and 38\% as offers contained actionable content, with many irrelevant tweets misclassified as requests or offers.

\subsection{Pre-trained encoder-only models}
MLMs based on BERT \cite{devlin2018bert} utilize only the encoder block of the transformer architecture \cite{vaswani2017attention}, making them ideal for tasks that require contextual embeddings. Several variations of BERT have been introduced, such as RoBERTa \cite{liu2019roberta}, MPNet \cite{song2020mpnet}, BERTweet \cite{nguyen2020bertweet}, and CrisisTransformers \cite{lamsal2024crisistransformers}. Such models have been applied to various tasks in crisis informatics, including classification of humanitarian content \cite{alam2021crisisbench}, identifying disaster-related \cite{prasad2023identification} and informative content \cite{alam2021crisisbench,koshy2023multimodal}, detecting location mentions \cite{suwaileh2023idrisi}, emotion classification \cite{myint2024unveiling}, stance detection \cite{poddar2022winds,cotfas2021longest,hayawi2022anti} and benchmarking \cite{lamsal2024crisistransformers}.

\subsubsection{Mini Models}
Model compression techniques, such as weight pruning \cite{han2015deep}, quantization \cite{gong2014compressing}, and knowledge distillation in a \textit{student-teacher network} \cite{hinton2015distillingknowledgeneuralnetwork}, aim to improve inference times and reduce model size. This study focuses on knowledge distillation, where a smaller student model mimics the output of a larger teacher model.

Several mini models, like DistilBERT \cite{sanh2020distilbertdistilledversionbert}, BERT$_{\text{medium/small/mini/tiny}}$ \cite{turc2019well}, PKD-BERT \cite{sun2019patient}, and TinyBERT \cite{jiao2019tinybert}, have been trained with varying sizes and configurations. However, these models remain general-purpose, i.e., they are not specifically tailored for crisis-related texts.

\section{Method}
Our proposed methodology consists of three key steps: (i) data labelling with generative LLMs in an ensemble, (ii) human validation of the labels, and (iii) designing mini models.

\subsection{The ``CrisisHelpOffer" Task}
\label{task}
The task is to classify social media texts shared during crises into four specific classes based on the nature and intent of the content: (i) \textit{Request}: Explicit requests for help, resources, or action; (ii) \textit{Offer}: Offers of assistance or resources; (iii) \textit{Irrelevant}: Tweets that do not request or offer material support, including emotional support, general information without action, or outdated requests; (iv) \textit{Request and Offer}: Tweets that simultaneously contain a request and an offer.

\subsection{Data Labelling and Human Validation}
\label{llms-ensemble}
Given a set of \( N \) tweets, \( T = \{t_1, t_2, \ldots, t_N\} \), we use four generative LLMs \( \mathcal{M} = \{\text{M}_1, \text{M}_2, \text{M}_3, \text{M}_4\} \). Considering the task definition, each model \( \text{M}_j \) classifies\footnote{A detailed prompt was designed for the task discussed in Section \ref{task}. The prompt will be shared with the dataset.} each tweet \( t_i \) into a label \( y_{i,j} \in \mathcal{Y} \), where:

\( \mathcal{Y} = \{\text{Request}, \text{Offer}, \text{Request and Offer}, \text{Irrelevant}\} \).

Next, we determine agreement across all models:
   \[
   C(t_i) = 
   \begin{cases} 
       1, & \text{if } y_{i,1} = y_{i,2} = y_{i,3} = y_{i,4}, \\
       0, & \text{otherwise}.
   \end{cases}
   \]

Now, we construct a labelled dataset \( T_{\text{agree}} \) by including only tweets for which all models agreed on the classification:

   \[
   T_{\text{agree}} = \{ t_i \in T : C(t_i) = 1 \}.
   \]

For each tweet in \( T_{\text{agree}} \), only the agreed-upon label is retained and utilized for subsequent human validation.

\subsubsection{Human Validation}
From \( T_{\text{agree}} \), we select a random sample (stratified) \( S \subset T_{\text{agree}} \) for human validation. The sample size \( |S| \) is based on a margin of error \( E = 3\% \) and a confidence level of \( 95\% \).

Each tweet \( t_i \in S \) is labelled by a human with a label \( y_{i,\text{human}} \) from the same label set \( \mathcal{Y} \). Next, we compute the kappa agreement \( \kappa \) to measure the consistency between LLM-assigned labels \( y_{i, \text{LLM}} \) and human-assigned labels \( y_{i, \text{human}} \) for all \( t_i \in S \). The value of \( \kappa \) determines the reliability of the LLM-generated labels in \( T_{\text{agree}} \), with higher values indicating strong agreement and thus greater confidence in the LLM labels as a substitute for human labelling.

\subsection{Mini Models for Crisis Texts}
\label{mini-autoencoders}

We select a teacher model \( T \) and design student models with smaller architectures (\textbf{M}edium, \textbf{S}mall and \textbf{T}iny), defined as follows:

\begin{centering}
S$_{\text{M}} = \{ \text{H}:512, \text{L}:8, \text{A}:8, \text{I}:2048\}$

S$_{\text{S}} = \{ \text{H}:384, \text{L}:6, \text{A}:6, \text{I}:1536 \}$

S$_{\text{T}} = \{ \text{H}:256, \text{L}:4, \text{A}:4, \text{I}:1024 \}$

\end{centering}

where, \textit{H} is the \textit{hidden size}, \textit{L} is the \textit{number of layers}, \textit{A} is the \textit{number of attention heads}, and \textit{I} is the \textit{intermediate size}.

\subsubsection{Knowledge Distillation}
We perform knowledge distillation \cite{hinton2015distillingknowledgeneuralnetwork} to create two sets of distilled models: (i) models distilled using soft labels (logits) and hard labels (discrete classes) of a \texttt{fine-tuned} $T$ on CrisisHelpOffer, and (ii) models distilled by approximating the embedding space of a \texttt{pre-trained} $T$. The former are \textit{task-specific models} (S$_{\text{i}}^{\text{T}}$), trained for the task defined in Section \ref{task}, while the latter are \textit{generic models} (S$_{\text{i}}^{\text{G}}$). We release S$_{\text{i}}^{\text{G}}$ for broader use in downstream classification tasks in crisis informatics. S$_{\text{i}}^{\text{T}}$ are presented for comparative purposes only, as their applicability is limited to the specific task defined in Section \ref{task}.

For S$_\text{i}^\text{T}$, we add a linear prediction layer on top with four output units, aligning with the number of classes in CrisisHelpOffer. For S$_{\text{i}}^{\text{G}}$, we define a linear downsampling network\footnote{This layer projects teacher embeddings to lower dimensions for each S$_{\text{i}}^{\text{G}}$.} \( D: \mathbb{R}^{d_T} \to \mathbb{R}^{d_S} \) where \( d_T \) and \( d_S \) are output dimensions of $T$ and S$_{\text{i}}^{\text{G}}$, respectively.

\subsubsection{Training Objective}
\label{losses}

To train S$_\text{i}^\text{T}$, we approximate the behaviour of $T$ using soft and hard labels. We use \textit{KL divergence} as the loss function for soft labels, while for hard labels, we use \textit{categorical cross-entropy} loss.

To train S$_{\text{i}}^{\text{G}}$, we minimize the Mean Squared Error (MSE):

\[
\mathbf{h}_{T,D} = D(\mathbf{h}_T)
\]
\[
\mathcal{L}_{\text{MSE}} = \frac{1}{n} \sum_{i=1}^{n} \|\mathbf{h}_{T,D} - \mathbf{h}_{\text{S}^\text{G}_\text{i}}\|^2
\]

where $\mathbf{h}_T = \text{teacher's embeddings}$ and $\mathbf{h}_{\text{S}^\text{G}_\text{i}} = \text{student's embeddings}$. This loss measures how well S$_{\text{i}}^{\text{G}}$ approximates the downsampled embeddings of $T$.

With respect to S$_{\text{i}}^{\text{G}}$, we experiment with updating students in two configurations: through mean-pooling of token embeddings and through $<$CLS$>$ token embedding.

\subsubsection{Optimization}
We optimize S$_\text{i}^\text{T}$ with a \textit{batch size} 32 and a \textit{learning rate} $2e-5$ until the F1 score saturates. We optimize S$_{\text{i}}^{\text{G}}$ for one epoch using \textit{mixed precisio}n on $\approx$440 million training samples, with a \textit{batch size} 1024 and a \textit{learning rate} $2e-4$.

\subsection{Fine-tuning}
\label{finetuning}
Following standard practice, we add a linear prediction layer to the output of an MLM \cite{nguyen2020bertweet}, using mean pooling over token embeddings. Fine-tuning configurations include a maximum of 30 epochs, a \textit{learning rate} $1e-5$, and a \textit{batch size} 32. Stratified sampling is used for generating train/validation/test splits (70/10/20) with \textit{train-test-split} from \textit{scikit-learn}, random state 42. F1 is used to assess performance after each epoch, with early stopping (patience 5, threshold 0.0001). Class weights are applied to address class imbalance. Fine-tuning is repeated 3 times, and final performance is reported as the average F1 (macro) at a 95\% confidence interval.

\subsection{Data}
\label{datasets}

\subsubsection{For training task-specific mini models (S$_\text{i}^\text{T}$)} \cite{lamsal2024crema} provides a dataset of 282k tweets, classified as requests or offers during a global crisis. The dataset contains a significant number of non-actionable or generic tweets unrelated to crisis situations. We refine this dataset by filtering non-actionable tweets as discussed in Section \ref{llms-ensemble}. We then fine-tune multiple MLMs on the refined dataset and use the best-performing classifier as $T$ to train S$_\text{i}^\text{T}$.

\subsubsection{For training generic mini models ( S$_{\text{i}}^{\text{G}}$)}
\label{large-scale-dataset}
We collected tweet identifiers from various publicly available tweet collections and hydrated them using Twitter's \textit{lookup endpoint} to generate a crisis text corpus containing $\approx$440 million tweets. The identifiers were sourced from CrisisNLP \cite{imran2016twitter}, DocNow catalog\footnote{https://catalog.docnow.io/}, and IEEE DataPort \cite{lamsal2023billioncov}. Below are some of the events captured in the dataset:

\begin{itemize}
    \item \textit{Natural Disasters}: Hurricanes Harvey, Irma, Florence, Dorian, Tropical Storm Imelda, Nepal Earthquake, Chile Earthquake, California Earthquake, Cyclone PAM, Typhoon Hagupit, India Floods, Pakistan Floods, Iceland Volcano, YMM Airport Fire 
    \item \textit{Conflicts, Wars, and Terrorism}: Israel-Palestine conflict, The fall of Aleppo, Las Ramblas Attack, Stockholm Attack 2017, Paris Attacks, Peshawar School Attack, 2017 Shooting in Las Vegas.
    \item \textit{Disease Outbreaks}: COVID-19 Pandemic, Middle East Respiratory Syndrome, Ebola Virus Outbreak
    \item \textit{Social and Political Protests}: \#J20 (activism, protests), Tyendinaga (protests, railway disruptions).
    \item \textit{Misc.}: Flight MH370, Climate Change \#PuertoRico
\end{itemize}

Each of these tweets is input to $T$ and S$_{\text{i}}^{\text{G}}$. The deviation of  S$_{\text{i}}^{\text{G}}$'s embedding from the downsampled embedding of $T$ is computed using the MSE loss, which is minimized, as discussed in Section \ref{losses}. This way, the embedding space of $T$ is approximated by S$_{\text{i}}^{\text{G}}$.

\subsubsection{Text Preprocessing}
We preprocess tweets as follows: URLs are replaced with ``HTTPURL" token, @mentions are replaced with ``@USER" token, HTML entities are decoded (e.g., \&amp; to \&), newline and multiple whitespaces are normalized to a single space, and emojis are replaced by their textual forms.

\section{Results and Discussion}

As discussed in Section \ref{llms-ensemble}, 282k tweets were labelled by four generative LLMs: Gemma 2 9B \cite{team2024gemma}, Llama 3.1 8B \cite{dubey2024llama}, and Ministral 8B and Mistral-Nemo 12B from \textit{Mistral AI}\footnote{https://mistral.ai/technology/}. Each LLM ran independently on 4 NVIDIA A100 80GB GPUs. The wall-clock times for labelling were\footnote{Such high wall-clock times make generative LLM-based approaches infeasible for real-time data classification scenarios with limited access to powerful computing resources.} (in days:hours-minutes): Gemma, \textit{2-23:31}; Llama, \textit{1-23:23}; Ministral, \textit{1-21:47}; and Mistral-Nemo, \textit{2-23:43}. The LLMs agreed on the same labels for 101k tweets (\( T_{\text{agree}} \)): 15.7k classified as ``Request", 5k as ``Offer", 38 as ``Request and Offer", and 80.5k as ``Irrelevant".

For human evaluation, we randomly sampled a stratified subset of 1057 tweets from \( T_{\text{agree}} \), plus all 38 tweets classified as ``Request and Offer" to avoid class under-representation, ensuring a 95\% confidence level and 3\% margin of error. Two independent annotators, both proficient in English, labelled tweets into four categories: ``Request", ``Offer", ``Request and Offer", and ``Irrelevant". Cohen's Kappa was 0.934 for \textit{LLMs vs. Human 1} and 0.924 for \textit{LLMs vs. Human 2}, indicating almost perfect agreement \cite{mchugh2012interrater}. Human evaluations show that the labels generated by the LLMs are of high quality, making the dataset reliable for training classifiers. Tweets in \( T_{\text{agree}} \) at this stage form the CrisisHelpOffer dataset.

\subsection{Classifier for CrisisHelpOffer}
\label{classifier-crisishelpoffer}

\begin{table}
\centering
\caption{Performance of different models on CrisisHelpOffer. The best F1 score is \textbf{highlighted}, and the second best is \underline{underlined}. For CrisisTransformers, the best model from each M1/M2/M3 variant is included.}
\label{finetuning-results}
\begin{tabular}{ll}
\toprule
Model & (avg. F1 ± std dev)                             \\
\midrule
BERT                    & 0.7611 ±0.0393 (±5.16\%)\\
RoBERTa                      & 0.7731 ±0.0403 (±5.21\%)\\
MPNet              & 0.7769 ±0.0463 (±5.96\%)\\
BERTweet & 0.7737 ±0.0441 (±5.70\%)\\
\midrule
\%CrisisTransformers\% & \\
CT-M1-Complete & \textbf{0.8485 ±0 (±0.00\%)}\\
CT-M2-OneLook  & 0.8066 ±0.0165 (±2.04\%)\\
CT-M3-OneLook  & \underline{0.8365 ±0.0236 (±2.82\%)}   \\
\midrule
S$^\text{T}_{\text{M}}$ & Soft:0.7893, Hard:0.7887\\
S$^\text{T}_{\text{S}}$ & Soft:0.7885, Hard:0.7879\\
S$^\text{T}_{\text{T}}$ & Soft:0.7845, Hard:0.7847\\
\bottomrule
\end{tabular}
\end{table}

As discussed in Section \ref{finetuning}, we fine-tuned multiple state-of-the-art MLMs on CrisisHelpOffer, including BERT \cite{devlin2018bert}, RoBERTa \cite{liu2019roberta}, MPNet \cite{song2020mpnet}, BERTweet \cite{nguyen2020bertweet}, and CrisisTransformers \cite{lamsal2024crisistransformers}. The fine-tuning results are summarized in Table \ref{finetuning-results}. Among all models, CrisisTransformers' \textit{CT-M1-Complete} achieved the highest F1 score of 0.8485 with zero variance\footnote{We evaluated the classifier on tweets from the Ukraine-Russia conflict to test CrisisHelpOffer's generalization. After inspecting 200 tweets (50 requests, 50 offers, and 100 irrelevant), we observed similar performance.}. In addition to its performance on CrisisHelpOffer, it has also shown strong performance across 18 different crisis classification tasks \cite{lamsal2024crisistransformers} compared to existing MLMs. Motivated by these results, we use \texttt{fine-tuned} and \texttt{pre-trained} versions of \textit{CT-M1-Complete} as teachers to train our mini models.

S$_\text{i}^\text{T}$ approximates the behaviour of the fine-tuned \textit{CT-M1-Complete}, and as a result, it remains task-specific to CrisisHelpOffer. In contrast, S$_{\text{i}}^{\text{G}}$, which approximates the embedding space of pre-trained \textit{CT-M1-Complete}, is a generic model suitable for any crisis classification task.

\subsection{Mini Models}
For both distillation types, three mini models with BERT architecture were initialized with random weights and trained independently on six A100 80GB GPUs. Table \ref{mini-models-arch} provides the architectural details. For S$_\text{i}^\text{T}$, we utilize the same train/test splits used to fine-tune MLMs in Section \ref{classifier-crisishelpoffer}. For S$_{\text{i}}^{\text{G}}$, the models were optimized in two configurations: the mean-pooled token embeddings or the $<$CLS$>$ token embedding for the student update. The teacher's embeddings were generated using mean-pooled token embeddings across both configurations. There were $\approx$440 million crisis-related tweets as training samples.

\begin{table}
\centering
\caption{Architecture of existing MLMs considered in this study and our mini models.}
\label{mini-models-arch}
\begin{tabular}{lllll|l}
\toprule
       & H & L & A & I & \#Params\\
\midrule
BERT   & 768         & 12     & 12    & 3072  & 109M\\
RoBERTa & 768         & 12     & 12    & 3072 & 125M\\
MPNet & 768         & 12     & 12    & 3072 & 133M \\
BERTweet & 768 & 12     & 12    & 3072 & 135M\\
CrisisTransformers & 768 & 12     & 12    & 3072 & 135M \\
\midrule
S$_{\text{M}}$ & 512         & 8      & 8     & 2048              & 58M \\
S$_{\text{S}}$  & 384         & 6      & 6     & 1536              & 35M \\
S$_{\text{T}}$   & 256         & 4      & 4     & 1024              & 19M\\
\bottomrule
\end{tabular}
\end{table}

Results show that all S$_\text{i}^\text{T}$ models converge to an F1 score of $\approx$0.78, regardless of their architecture or parameter count, in both soft and hard label configurations (refer to Table \ref{finetuning-results}). These models outperform the existing MLMs.

\begin{figure}
    \centering
\includegraphics[width=0.60\linewidth]{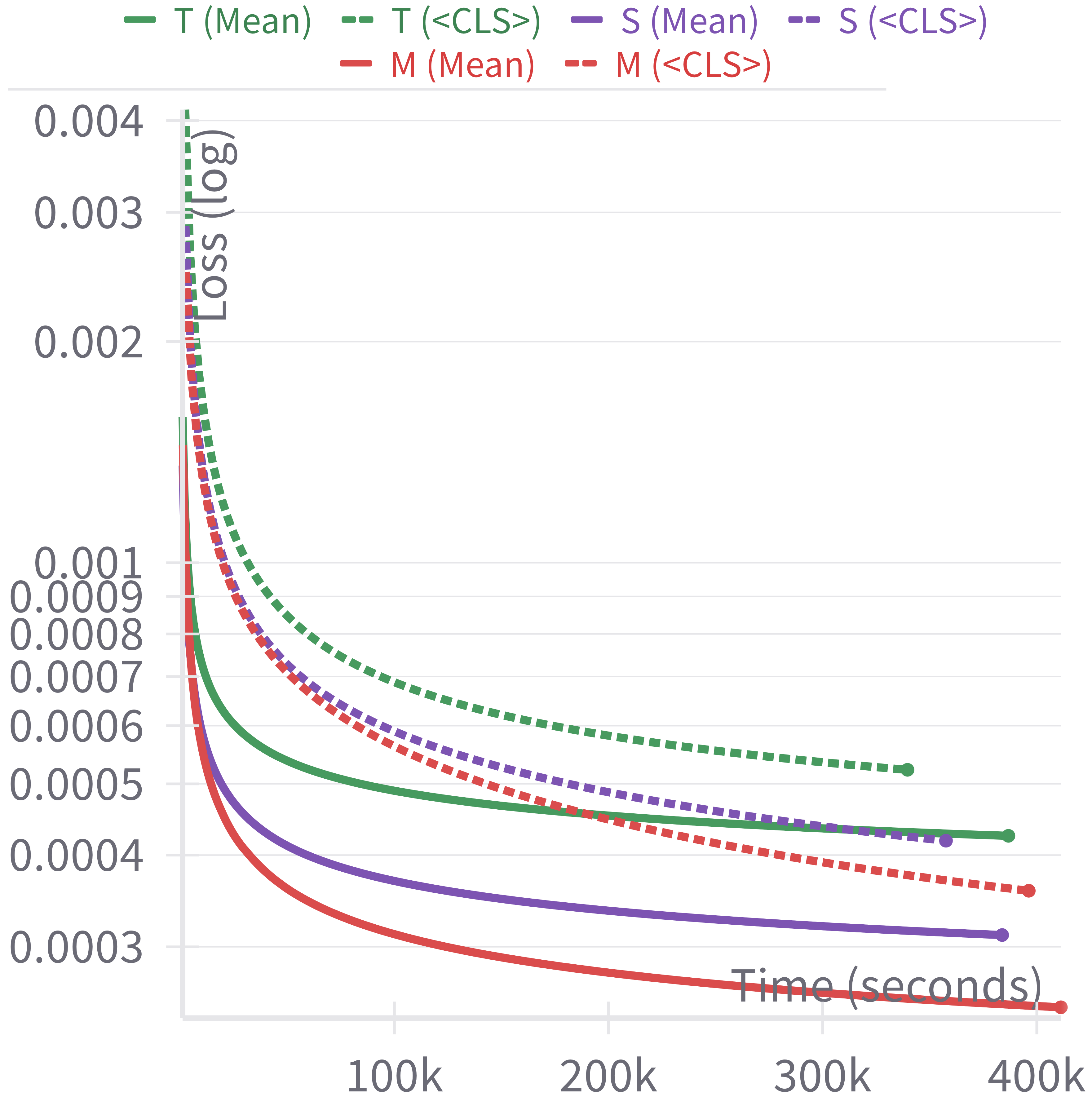}
    \caption{Loss curves ($y$-axis is in log scale) for S$_{\text{i}}^{\text{G}}$ (T: S$_\text{T}^\text{G}$, S: S$_\text{S}^\text{G}$ and M: S$_\text{M}^\text{G}$). During training, the teacher’s token embeddings were mean-pooled. The difference in training was in how students were updated: (i) mean-pooling of tokens and (ii) $<$CLS$>$ token. All variants started converging after 75\% of the total steps.}
    \label{loss-curves}
\end{figure}

S$_{\text{i}}^{\text{G}}$ finished training in 4-5 days. The resulting loss curves from the training are shown in Figure \ref{loss-curves}. Results suggest that the mean-pooled approach provides more effective training signals for the students, resulting in lower loss values and, potentially, better performance on downstream tasks. Therefore, we consider only the mean-pooled models for the evaluation.

To evaluate the robustness of S$_{\text{i}}^{\text{G}}$, we fine-tuned them on CrisisHelpOffer and 12 additional human-labelled crisis classification datasets. These datasets include CrisisLex \cite{olteanu2014crisislex}, CrisisNLP \cite{imran2016twitter}, COVID-19 Stance \cite{poddar2022winds}, Stress-annotated datasets \cite{mauriello2021sad}, LocBERT \cite{lamsal2022did}, HMC \cite{biddle2020leveraging}, Vax Opinions \cite{cotfas2021longest}, PHM \cite{karisani2018did}, and ANTiVax \cite{hayawi2022anti}. Existing MLMs~---~BERT, RoBERTa, MPNet, BERTweet~---~and S$_{\text{i}}^{\text{G}}$ were fine-tuned under same settings, as discussed in Section \ref{finetuning} on these 13 crisis classification tasks. The results are summarized in Table \ref{finetuning-results-mini-autoencoders}.

Against task-specific counterparts, on CrisisHelpOffer, S$_\text{M}^\text{G}$ achieves a 4.43\% improvement, but S$_\text{M}^\text{G}$ and S$_\text{T}^\text{G}$ underperform by 2.44\% and 7.67\%, respectively.

\begin{table*}
\centering
\caption{Performance of existing MLMs and S$_{\text{i}}^{\text{G}}$ on CrisisHelpOffer and 12 additional crisis classification tasks.}
\label{finetuning-results-mini-autoencoders}
\begin{tabular}{l|llll|lll}
\toprule
& \multicolumn{4}{c|}{SOTA encoder-only models} & \multicolumn{3}{c}{(\textbf{our}) mini models}  \\
\midrule
Task $\downarrow$                                     & BERT   & RoBERTa & MPNet  & BERTweet & S$_\text{M}^\text{G}$ & S$_\text{S}^\text{G}$ & S$_\text{T}^\text{G}$ \\
\midrule
CrisisHelpOffer                           & 0.7611 & 0.7731  & 0.7769 & 0.7737   & 0.8243 (±3.50\%) & 0.7693 (±4.89\%) & 0.7245 (±2.18\%) \\
\midrule
CrisisLex                                 & 0.7297 & 0.7603  & 0.7571 & 0.7569   & 0.7504 (±0.14\%) & 0.7417 (±0.67\%) & 0.7252 (±0.07\%) \\
CrisisNLP                                 & 0.7782 & 0.7919  & 0.7776 & 0.7871   & 0.7721 (±1.49\%) & 0.7080 (±3.30\%) & 0.6981 (±1.60\%) \\
\citeauthor{poddar2022winds} & 0.5197 & 0.5808  & 0.5810  & 0.5890    & 0.6463 (±0.53\%) & 0.6222 (±1.20\%) & 0.5933 (±0.27\%) \\
SAD COVID                                 & 0.9001 & 0.9125  & 0.9208 & 0.9358   & 0.9051 (±0.00\%) & 0.9444 (±0.13\%) & 0.9275 (±0.22\%) \\
SAD Stress                                & 0.6770 & 0.7122  & 0.6758 & 0.6940    & 0.6841 (±3.12\%) & 0.6838 (±2.10\%) & 0.6371 (±1.80\%) \\
SAD Stressor                              & 0.7040 & 0.7171  & 0.6934 & 0.7148   & 0.6677 (±0.75\%) & 0.6551 (±0.16\%) & 0.6406 (±0.34\%) \\
LocBERT                                   & 0.7230 & 0.7665  & 0.7590  & 0.7727   & 0.7893 (±0.12\%) & 0.7634 (±1.10\%) & 0.7294 (±1.40\%) \\
HMC (a)                                   & 0.8745 & 0.8904  & 0.8882 & 0.9009   & 0.8871 (±0.31\%) & 0.8863 (±0.15\%) & 0.8677 (±0.72\%) \\
HMC (b)                                   & 0.9899 & 0.9927  & 0.9905 & 0.9933   & 0.9932 (±0.04\%) & 0.9922 (±0.04\%) & 0.9911 (±0.02\%) \\
Vax Opinions                        & 0.7595 & 0.8338  & 0.8119 & 0.8562   & 0.8663 (±0.32\%) & 0.8396 (±0.49\%) & 0.8236 (±0.63\%) \\
PHM                                       & 0.8106 & 0.831   & 0.8029 & 0.8209   & 0.8186 (±0.52\%) & 0.8074 (±0.00\%) & 0.7805 (±0.71\%) \\
Anti-Vax                                  & 0.9748 & 0.9837  & 0.9829 & 0.9830    & 0.9859 (±0.05\%) & 0.9829 (±0.07\%) & 0.9776 (±0.10\%) \\
\midrule
Macro Avg.                                & 0.785  & 0.811   & 0.8    & 0.813    & 0.815            & 0.799            & 0.778       \\
\bottomrule
\end{tabular}
\end{table*}

\begin{table}
\centering
\caption{Evaluations of selected existing mini MLMs on tasks listed in Table \ref{finetuning-results-mini-autoencoders}. Due to space constraints, we provide a macro average only for these models.}
\label{mini-models-results}
\begin{tabular}{l|llll|l}
\toprule
                & \multicolumn{4}{c|}{Architecture} &            \\
                \midrule
                & H       & L    & A     & I       & Macro Avg. \\
\midrule
DistilBERT      & 768   & 6  & 12  & 3072  & 0.761      \\
BERT$_{\text{medium}}$ & 512   & 8  & 8   & 2048  & 0.758      \\
BERT$_{\text{small}}$  & 512   & 4  & 8   & 2048  & 0.757      \\
BERT$_{\text{mini}}$   & 256   & 4  & 4   & 1024  & 0.728      \\
TinyBERT$_{4}$  & 312   & 4  & 12  & 1200  & 0.753      \\
\bottomrule
\end{tabular}
\end{table}

Overall, S$_\text{M}^\text{G}$, S$_\text{S}^\text{G}$, and S$_\text{T}^\text{G}$ achieve competitive performance, often surpassing or closely matching the base architectures of BERT, RoBERTa, MPNet and BERTweet. In this section, we use BERT as our baseline for discussion, as the literature predominantly features distilled versions of BERT. Out of 13 datasets, S$_\text{M}^\text{G}$ outperformed the baseline on 11 datasets, S$_\text{S}^\text{G}$ on 10 datasets, and S$_\text{T}^\text{G}$ on 5 datasets. For S$_\text{T}^\text{G}$, the performance drop relative to the baseline was less than 5\% across 4 datasets. On average, S$_\text{M}^\text{G}$ and S$_\text{S}^\text{G}$ achieved performance improvements of 3.82\% and 1.78\%, respectively, while S$_\text{T}^\text{G}$ showed a decrease of 0.89\% compared to the baseline. To benchmark against existing mini MLMs, we evaluated DistilBERT, BERT$_{\text{medium}}$, BERT$_{\text{small}}$, BERT$_{\text{mini}}$, and TinyBERT$_4$, each with varying configurations of hidden size (H), layers (L), attention heads (A), and intermediate size (I). These existing mini MLMs were fine-tuned on the same 13 datasets using the training setup detailed in Section \ref{finetuning}. The results, summarized in Table \ref{mini-models-results}, show that all existing mini MLMs underperform relative to the baseline: DistilBERT by -3.06\%, BERT$_{\text{medium}}$ by -3.44\%, BERT$_{\text{small}}$ by -3.57\%, BERT$_{\text{mini}}$ by -7.26\%, and TinyBERT$_4$ by -4.08\%.

Regarding inference times, our models S$_{i}$ offer substantial speedup advantages over the baseline. Inference was conducted on an NVIDIA A100 GPU with a batch size of 32, over 1,000 iterations, and a few initial warm-up passes to stabilize performance. Results are summarized in Table \ref{inference-times}. S$_{\text{M}}$, S$_{\text{S}}$, and S$_{\text{T}}$ achieve throughput improvements of 3.5x, 7.7x, and 18.6x, respectively. Our mini models have $\leq$ attention heads, hidden sizes, and intermediate sizes compared to their counterparts with the same $L$. The throughput improvements hold even when compared to existing mini MLMs. These gains make S$_{i}$ more suitable for real-time or high-throughput crisis classification tasks.

\begin{table}
\centering
\caption{
Inference times on an NVIDIA A100 GPU. Batch$_{32}$ denotes seconds per batch (size = 32), Throughput is \textit{samples/sec}, and $\Delta_{\text{baseline}}$ is the throughput improvement factor over BERT. Models are grouped by layer count ($L$).}
\label{inference-times}
\begin{tabular}{p{2.5cm}p{1.4cm}p{1.4cm}p{1.4cm}}
\toprule
           & Batch$_{32}$  & Throughput & $\Delta_{\text{baseline}}$ \\
\midrule
\%base arch.\% & & & \\
BERT       & 0.0305 & 1,050      & -     \\
RoBERTa & 0.0304 & 1,051 & -  \\
MPNet &  0.0277 & 1,155 & x1.1\\
BERTweet &  0.0261 & 1,224 & x1.16\\
CrisisTransformers & 0.0251 & 1275 & x1.21\\
\midrule
S$_{M}$ (\textbf{our})    & 0.0087 & 3,699      & x3.5  \\
BERT$_{\text{medium}}$ & 0.0088 & 3,634      & x3.5  \\
\midrule
S$_{S}$ (\textbf{our})    & 0.0039 & 8,160      & x7.7  \\
DistilBERT & 0.016  & 2,002      & x1.9  \\
\midrule
S$_{T}$ (\textbf{our})      & 0.0016 & 19,549     & x18.6 \\
BERT$_{\text{small}}$  & 0.0045 & 7,164      & x6.8  \\
BERT$_{\text{mini}}$   & 0.0017 & 18,610     & x17.7 \\
TinyBERT$_{4}$   & 0.0027 & 11,779     & x11.2 \\

\bottomrule
\end{tabular}
\end{table}

\section{Case Study}
This section presents a spatiotemporal analysis of a global crisis event to explore the distribution of requests and offers across selected countries, examining regional responses to seeking or providing assistance. We used \textit{MegaGeoCOV Extended} \cite{lamsal2023twitter}, which contains 17.8 million EN tweets from the COVID-19 pandemic (10/2019 to 10/2022) with geographic data. These tweets were classified with the \textit{CT-M1-Complete} classifier from Table \ref{finetuning-results}, resulting in 303.7k Request tweets and 186.6k Offer tweets for analysis. Table \ref{top-10-countries-r/o} presents tweet distributions and requests-to-offers (R/O) ratios for the top 10 countries, while Table \ref{top-10-cities} lists the top cities by tweet volume.

\subsection{Key Findings}
Countries such as India, South Africa, and Pakistan exhibited a high demand for help. The United States had the most balanced R/O ratio, while Ireland and Canada had more offers than requests. The overall trend indicates that developing countries tend to have a higher demand for requests than offers, whereas developed countries tend to show more balanced or offer-dominant behaviour. These findings are further supported by Table \ref{top-10-cities}, which reveals that 9 Indian cities rank among the top cities posting request tweets, compared to only 2 Indian cities appearing on the list for offers.

\begin{figure}
    \centering
    \includegraphics[width=0.49\textwidth]{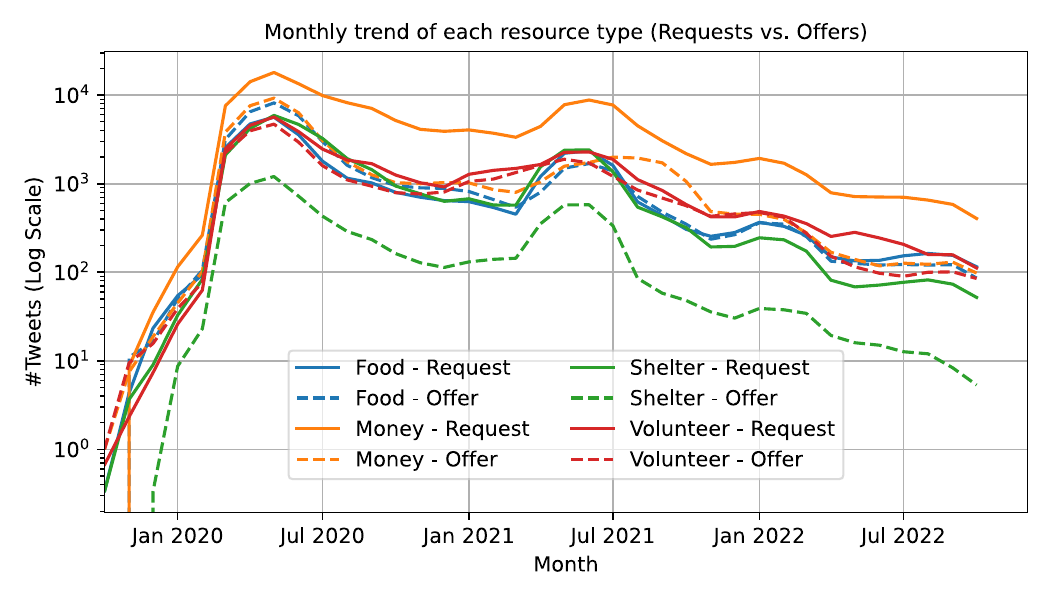}
    \caption{Distribution of request and offers tweets across resource types: Food, Money, Shelter, and Volunteer.}
    \label{global-trend-resources}
\end{figure}

We fine-tuned the MLMs listed in Table \ref{finetuning-results} on a dataset from \cite{purohit2014emergency}, which includes labelled crisis tweets across six resource types: money, volunteers, clothing, shelter, medical aid, and food. CrisisTransformers' \textit{CT-M1-Complete} achieved the best performance (F1: 0.9809, Precision: 0.9834, Recall: 0.9787). We used this classifier to categorize request and offer tweets into these resource types. We manually evaluated how it handled tweets related to PPEs and face masks, which were classified under clothing, as the dataset did not account for virus outbreaks. Thus, we focused on tweets related to Food, Money, Shelter, and Volunteers. Figure \ref{global-trend-resources} shows the monthly trend of requests and offers globally across these four resources. Requests for these resources were minimal before 2020 but surged starting early that year, peaking between March and May. Money was the most requested resource (50\%), followed by volunteers (14\%), shelter (12\%), and food (11\%). Offers followed a similar pattern, with money (28\%), food (23\%), volunteers (18\%), and shelter (4\%) being the most frequently offered. From mid-2021 onward, both requests and offers declined, though they remained above pre-pandemic levels.

\begin{figure}
    \centering
\includegraphics[width=0.49\textwidth]{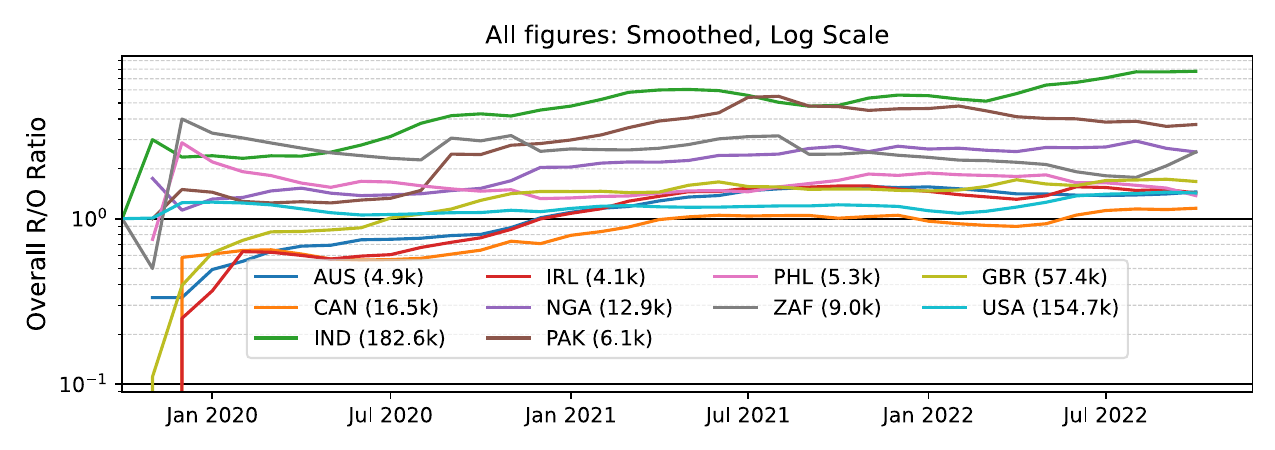}
    \caption{R/O distributions across selected countries.}
    \label{global-trend-R/O}
\end{figure}

Next, we explored the temporal trends of R/O ratios, focusing on the top 10 countries, both overall (Figure \ref{global-trend-R/O}) and by resource type: money, volunteer, and shelter. Results show that developing countries consistently have higher R/O ratios, with India leading across all resource types, followed by Nigeria and Pakistan. In contrast, developed countries like the USA, Canada, and the UK tend to have more balanced or offer-dominant dynamics, with R/O ratios near or below 1. Resource-specific trends provide further insights: for ``Money", R/O ratios gradually increase in many countries, with India and Pakistan exceeding 10. For ``Shelter", India consistently shows high R/O ratios, surpassing 10, especially after early 2021, while Nigeria and Pakistan also show upward trends with some fluctuations. For ``Volunteer", India and Pakistan maintain high ratios, with Pakistan spiking in 2020 before stabilizing in mid-2021. Developed nations generally have R/O ratios below 1. These trends highlight disparities in crisis-related needs and resource availability, with developing nations facing greater imbalances due to more requests than offers.

\begin{table}
\centering
\caption{Distributions of tweets and R/O for top 10 countries.}
\label{top-10-countries-r/o}
\begin{tabular}{p{3.5cm}p{1.1cm}p{1cm}p{1.1cm}}
\toprule
Country (ISO a-3)                     & \#requests & \#offers & R/O ($\downarrow$)      \\
\midrule
India (IND)                       & 143.3k     & 39.2k    & 3.65 \\
South Africa (ZAF)                & 5.8k       & 3.1k     & 1.88 \\
Pakistan (PAK)                    & 3.9k       & 2.1k     & 1.85 \\
Nigeria (NGA)                     & 8k       & 4.8k     & 1.66 \\
Philippines (PHL) &                 3k       & 2.3k     & 1.29 \\
United Kingdom (GBR)              & 32.1k      & 25.1k    & 1.27 \\
Australia (AUS)                   & 2.6k       & 2.2k     & 1.17 \\
United States (USA)               & 76.8k      & 77.8k    & 0.99 \\
Ireland (IRL)                     & 19k       & 2.2k     & 0.84 \\
Canada (CAN)                      & 6.8k       & 9.6k     & 0.72 \\
\bottomrule
\end{tabular}
\end{table}

\begin{table}
\centering
\caption{Top cities sorted by total volume of tweets.}
\label{top-10-cities}
\begin{tabular}{ll|ll}
\toprule
\multicolumn{2}{c}{Request}       & \multicolumn{2}{c}{Offer}   \\
\midrule
City                   & \% ($\downarrow$)      & City             & \% ($\downarrow$)      \\
\midrule
New Delhi, India       & 4.63* & New Delhi, India & 2.19 \\
Mumbai, India          & 4.31 & Mumbai, India    & 1.70 \\
Bengaluru S., India & 1.28 & Los Angeles, CA  & 1.27 \\
Kolkata, India         & 1.24 & Manhattan, NY    & 1.06 \\
Gurgaon, India         & 0.97 & Toronto, Ontario & 0.98 \\
Hyderabad, India       & 0.94 & Washington, DC   & 0.77 \\
Noida, India           & 0.87 & Houston, TX      & 0.75 \\
Los Angeles, CA        & 0.75 & Chicago, IL      & 0.75 \\
Bengaluru, India       & 0.75 & Lagos, Nigeria   & 0.71 \\
Haveli, India          & 0.66 & Brooklyn, NY     & 0.62\\
\bottomrule
\end{tabular}
\end{table}

\section{Conclusion and Future Directions}
This study introduced CrisisHelpOffer, a dataset comprising 101k labelled tweets designed for training classifiers to identify actionable requests and offers shared on social media during crises. Additionally, we presented the first set of \textit{mini models} for the crisis informatics domain, establishing new benchmarks in the field. These models maintain a balance of contextual understanding of crisis-related texts and computational efficiency, making them well-suited for real-time or high-throughput crisis classification tasks. Lastly, we analyzed tweets on the COVID-19 pandemic, uncovering disparities in crisis-related needs and resource availability between developing and developed nations.

Future research could expand CrisisHelpOffer by incorporating multi-lingual tweets and data from diverse platforms, improving its global relevance. Developing multi-lingual mini models for the crisis domain can be another research avenue. Furthermore, integrating volumetric patterns of help requests into early warning systems could improve situational awareness for humanitarian organizations.

\section*{Ethics Statement}
\label{ethical-con}
The collection of tweets in this study was done in compliance with Twitter’s terms of use. To protect individuals' privacy, we masked tweet contents (e.g., usernames, profile links) that could identify individuals before using them for LLM processing, human validation, knowledge distillation, fine-tuning or spatiotemporal analysis.

The pretraining corpus of CrisisTransformers had unfiltered tweets, resulting in a significant volume of non-neutral content. Since our mini models approximate the embedding space of a teacher model from this family, both the mini models and their fine-tuned versions can, as is the case with any MLMs, produce biased predictions, showing partiality toward certain groups, perspectives, or sentiments present in the training data.

\bibliographystyle{named}
\bibliography{bib}

\end{document}